\documentclass[twoside,11pt]{article}

\usepackage{array}
\usepackage{booktabs}
\usepackage{makecell}
\usepackage{lscape}
\usepackage{tabularx}

\usepackage{bm}
\usepackage{amssymb}
\usepackage{amsmath}
\usepackage{mathtools}
\usepackage{graphicx}
\usepackage{subcaption}
\usepackage{natbib}
\usepackage{jmlr2e}

\usepackage{hyperref}
\hypersetup{
  colorlinks=false,
}

\newcommand{\dataset}{{\cal D}}

\newcommand{\dsname}[1]{\texttt{#1}}
\newcommand{\package}[1]{\texttt{#1}}

\newcommand{\bs}[1]{\boldsymbol{#1}}
\newcommand{\ts}{\textsuperscript}

\newcommand{\bigo}[1]{\mathcal{O}\left(#1\right)}

\newcommand{\errout}{\mathrm{Err}_{\mathrm{out}}}

\newcommand{\twodot}[1]{\ddot{#1}}
\newcommand{\threedot}[1]{\dddot{#1}}
\newcommand{\fourdot}[1]{\ddddot{#1}}

\newcommand{\threelu}{\dddot{\ell}_i\left(u\right)}
\newcommand{\fourlu}{\ddddot{\ell}_i\left(u\right)}

\newcommand{\utildei}{\tilde{u}_{/i}}

\newcommand{\onelui}{\dot{\ell}_i\left(u_i\right)}
\newcommand{\twolui}{\ddot{\ell}_i\left(u_i\right)}
\newcommand{\threelui}{\dddot{\ell}_i\left(u_i\right)}
\newcommand{\fourlui}{\ddddot{\ell}_i\left(u_i\right)}

\newcommand{\onelutildei}{\dot{\ell}_i(\utildei)}
\newcommand{\twolutildei}{\ddot{\ell}_i(\utildei)}

\newcommand{\utildedef}{u_i + \frac{\onelui h_i}{1 - \twolui h_i}}

\newcommand{\betahat}{\bs{\hat{\beta}}}

\newcommand{\reg}{R_{\bs{\lambda}}}

\newcommand{\tworeg}{\ddot{r}_{j, \bs{\lambda}}}
\newcommand{\threereg}{\dddot{r}_{j, \bs{\lambda}}}

\newcommand{\regj}{r_{j, \bs{\lambda}}(\hat{\beta}_j)}

\newcommand{\threeregj}{\dddot{r}_{j, \bs{\lambda}}(\hat{\beta}_j)}
\newcommand{\fourregj}{\ddddot{r}_{j, \bs{\lambda}}(\hat{\beta}_j)}

\newcommand{\ldivexpr}{1 - \twolui h_i}

\ShortHeadings{Optimizing Approximate Leave-one-out Cross-Validation to Tune Hyperparameters}{Burn}
\firstpageno{1}

\begin{document}

\title{Optimizing Approximate Leave-one-out Cross-validation to Tune Hyperparameters}

\author{\name Ryan Burn \email ryan.burn@gmail.com \\
       \addr Department of Mathematics\\
       University of Washington\\
       Seattle, WA 98195-4350, USA}

\editor{}

\maketitle

\begin{abstract}
For a large class of regularized models, leave-one-out cross-validation can be efficiently
estimated with an approximate leave-one-out formula (ALO). We consider the problem of adjusting
hyperparameters so as to optimize ALO. We derive efficient formulas to compute the
gradient and hessian of ALO and show how to apply a second-order optimizer to find hyperparameters.
We demonstrate the usefulness of the proposed approach by finding hyperparameters for 
regularized logistic regression and ridge regression on various real-world data sets.
\end{abstract}


\section{Introduction}

Let $\dataset = \{\left(y_1, \bs{x}_1\right),\cdots,\left(y_n, \bs{x}_n\right)\}$ denote
a data set, where $\bs{x}_i \in \mathbb{R}^p$ are features and $y_i \in \mathbb{R}$ are
responses. In many applications, we model the observations as independent and identically distributed
draws from an unknown joint distribution of the form
$q\left(y_i, \bs{x}_i\right)=q_1\left(y_i \mid \bs{x}_i^\top \bs{\beta}_*\right) q_2\left(\bs{\beta}_*\right)$,
and we estimate $\bs{\beta}_*$ using the optimization problem
\begin{equation}\label{subobj}
  \betahat \triangleq     \operatorname*{argmin}_{\bs{\beta}} \left\{ 
          \sum_{i=1}^n \ell\left(y_i \mid \bs{x}_i^\top \bs{\beta}\right) +
          \reg\left(\bs{\beta}\right)
      \right\},
\end{equation}
where $\ell$ is a loss function, $\reg(\bs{\beta}) = \sum_{j=1}^p \regj$ is a regularizer, 
and $\bs{\lambda}$ represents hyperparameters.
$\reg$ controls the complexity of the model by penalizing larger values of $\bs{\beta}$ and can 
prevent overfitting. We aim to select $\bs{\lambda}$ so as to minimize the out-of-sample prediction 
error
\begin{align*}
  \errout \triangleq \mathbb{E}\left[
    \ell\left(y_o\mid\bs{x}_o^\top \betahat\right) \mid \dataset
    \right],
\end{align*}
where $\left(y_o,\bs{x}_o\right)$ is an unseen sample from the distribution $q\left(y,\bs{x}\right)$.
Because $q$ is unknown, we estimate $\errout$ with a function 
$f\left(\bs{\lambda}\mid \dataset\right)$ and apply a hyperparameter optimization algorithm to find
\begin{align*}
\bs{\hat{\lambda}} \triangleq \operatorname*{argmin}_{\bs{\lambda}} 
    f\left(\bs{\lambda}\mid\dataset\right).
\end{align*}
Empirical evidence has shown leave-one-out cross-validation (LO) to be an accurate method for
estimating $\errout$ \citep{rad2:20}. In general, LO can be expensive to compute,
requiring a model to be fit $n$ times; however, under certain conditions, it can be efficiently 
estimated using a closed-form approximate leave-one-out formula (ALO), and the approximation has 
been shown to be accurate in high-dimensional settings \citep{rad:20}. In addition, for the special 
case of ridge regression, ALO is exact and equivalent to Allen's PRESS \citep{allen:74}. 

In this paper, we derive efficient formulas to compute the gradient and hessian of ALO, given a
smooth loss function and a smooth regularizer, and we show how to apply a trust-region algorithm
to find hyperparameters for several different classes of regularized models.

\subsection{Relevant Work}
Grid search is a commonly used approach for hyperparameter optimization.
Although it can work well for models with only a single hyperparameter, it
requires a search space to be specified and quickly becomes inefficient when
multiple hyperparameters are used \citep{bergstra:12}. Other scalable
gradient-based approaches have focused on using holdout sets to approximate
$\mathrm{Err}_{\mathrm{out}}$ \citep{do:08,bengio:00}. 
Using ALO, we expect our objective function to be a more accurate
proxy for $\mathrm{Err}_{\mathrm{out}}$. We further improve on previous
approaches by computing the hessian of our objective function and applying
a trust-region algorithm, allowing us to make global convergence guarantees.

\subsection{Notation}
We denote vectors by lowercase bold letters and matrices by uppercase bold letters. We use the notation
$\mathrm{vec}\left[\left\{a_i\right\}_i\right]$ to denote the column vector 
$\left(a_1, a_2, \ldots\right)^\top$. For a function
$f\colon \mathbb{R} \to \mathbb{R}$, we denote its 1\ts{st}, 2\ts{nd}, 3\ts{rd}, and 4\ts{th} 
derivatives by $\dot{f}$, $\twodot{f}$, $\threedot{f}$, and $\fourdot{f}$, respectively. 
Finally, given a function $f\colon \mathbb{R}^p \to \mathbb{R}$, we denote the gradient by 
$\nabla f$ and the hessian by $\nabla^2 f$.

\section{Preliminary: Trust-region Methods}
Let $f\colon \mathbb{R}^p \to \mathbb{R}$ denote a twice-differentiable objective function. 
Trust-region methods are iterative, second-order optimization algorithms that produce a sequence
$\left\{\bs{x}_k\right\}$, where the $k$\ts{th} iteration is generated by updating the previous iteration
with a solution to the subproblem \citep{sorensen:82}
\begin{align*}
  \bs{x}_k &= \bs{x}_{k-1} + \bs{\hat{s}}_k \quad\mbox{and} \\
  \bs{\hat{s}}_k &= \operatorname*{argmin}_{\bs{s}} \left\{
      f\left(\bs{x}_{k-1}\right) +
      \nabla f\left(\bs{x}_{k-1}\right)^\top \bs{s} + 
      \frac{1}{2} \bs{s}^\top \nabla^2 f\left(\bs{x}_{k-1}\right) \bs{s} 
     \right\} \\
   &\phantom{=} \quad\mathrm{s.t.} \quad \| \bs{s} \| \le \delta_k.
\end{align*}
The subproblem minimizes the second-order approximation of $f$ at $\bs{x}_{k-1}$ within the 
neighborhood $\| \bs{s} \| \le \delta_k$, called the trust region. Using the trust region, 
we can restrict the second-order approximation to areas where it models $f$ well. Efficient algorithms
exist to solve the subproblem regardless of whether $\nabla^2 f(\bs{x}_{k-1})$ is positive-definite,
making trust-region methods well-suited for non-convex optimization problems \citep{more:83}. 
With proper rules for updating $\delta_k$ and standard assumptions, such as Lipschitz continuity of 
$\nabla f$, trust-region methods are globally convergent. Moreover, if $\nabla^2 f$ is 
Lipschitz continuous for all $\bs{x}$ sufficiently close to a nondegenerate second-order stationary 
point $\bs{x}_*$, where $\nabla^2 f(\bs{x}_*)$  is positive-definite, then trust-region methods 
have quadratic local convergence \citep{nocedal:99}.

\section{Preliminary: ALO}
Put $\ell_i(u) \triangleq \ell(y_i \mid u)$.  The LO estimate for $\mathrm{Err}_{\mathrm{out}}$ is 
defined as
\begin{align*}
  \mathrm{LO}_{\bs{\lambda}} \triangleq \frac{1}{n} 
    \sum_{i=1}^n \ell_i\left(\bs{x}_i^\top \betahat_{/i}\right),
\end{align*}
where
\begin{equation}\label{subobjminus}
  \betahat_{/i} \triangleq     \operatorname*{argmin}_{\bs{\beta}} \left\{ 
          \sum_{j \ne i} \ell_j\left(\bs{x}_j^\top \bs{\beta}\right) +
          \reg\left(\bs{\beta}\right)\right\}.
\end{equation}
ALO works by finding $\betahat$ and then using a single step of Newton's method to approximate 
$\betahat_{/i}$ from the gradient and hessian of Equation~\ref{subobjminus} at $\betahat$. 
Let $\bs{g}_{/i}$ and $\bs{H}_{/i}$ denote the gradient and hessian, respectively, of 
Equation~\ref{subobjminus} at $\betahat$. The ALO approximation to 
$\bs{\hat{\beta}}_{/i}$ is 
$\bs{\tilde{\beta}}_{/i} \triangleq \betahat - \bs{H}_{/i}^{-1} \bs{g}_{/i}$,
and
\begin{align*}
  \mathrm{ALO}_{\bs{\lambda}} \triangleq \frac{1}{n}\sum_{i=1}^n \ell_i\left(
    \bs{x}_i^\top \bs{\tilde{\beta}}_{/i}
   \right).
\end{align*}
Computed naively, this formula would require solving a linear system of order $p$ $n$ times, but
we can achieve a more efficient form by applying the matrix inversion lemma. Let $\bs{X}$ represent
the feature matrix whose i\ts{th} row is $\bs{x}_i$; let $\bs{H}$ denote the hessian of
Equation~\ref{subobj} at $\betahat$. Define 
$\bs{W} \triangleq \nabla^2 \reg(\betahat)$ and $u_i = \bs{x}_i \betahat$.
Then,
\begin{align*}
\bs{H} = \bs{X}^\top \bs{A} \bs{X} + \bs{W},
\end{align*}
where $\bs{A}$ is a diagonal matrix with $\bs{A}_{ii} = \twodot{\ell_i}\left(u_i\right)$, and
\begin{align*}
\mathrm{ALO}_{\bs{\lambda}} = \frac{1}{n} \sum_{i=1}^n \ell_i\left(\utildedef\right),
\end{align*}
where $h_i \triangleq \bs{x}_i^\top \bs{H}^{-1} \bs{x}_i$. See \citet{rad:20} for a derivation.

\section{Optimizing ALO}
Put
\begin{align*}
  \utildei \triangleq \utildedef
  \quad\mbox{and}\quad
  f\left(\bs{\lambda}\right)  \triangleq \frac{1}{n}\sum_{i=1}^n \ell_i(\utildei).
\end{align*}
We present formulas 
for computing $\nabla f$ and $\nabla^2 f$. See Appendix A and Appendix B for
derivations. In general, $\nabla^2 f$ will not be positive-definite; however, we can use a 
trust-region algorithm to find local minimums.
\begin{theorem}
\label{alogradient}
The gradient of $\mathrm{ALO}_{\bs{\lambda}}$ can be computed as
\begin{align*}
\frac{\partial f}{\partial \lambda_s} &= \frac{1}{n}\sum_{i=1}^n\left(
      \onelutildei \times \frac{\partial \utildei}{\partial \lambda_s}\right),
\end{align*}
where
\begin{align*}
  \frac{\partial \utildei}{\partial \lambda_s} &=
    \frac{\partial \utildei}{\partial u_i} \times \frac{\partial u_i}{\partial \lambda_s} +
    \frac{\partial \utildei}{\partial h_i} \times \frac{\partial h_i}{\partial \lambda_s}, \\
  \frac{\partial u_i}{\partial \lambda_s} &= 
      \bs{x}_i^\top \frac{\partial \betahat}{\partial \lambda_s}, \\
  \frac{\partial \betahat}{\partial \lambda_s} &=
      -\bs{H}^{-1}\frac{\partial \nabla \reg}{\partial \lambda_s}(\betahat), \\
  \frac{\partial h_i}{\partial \lambda_s} &= 
      -\bs{x}_i^\top \bs{H}^{-1} 
            \frac{\partial \bs{H}}{\partial \lambda_s} \bs{H}^{-1} \bs{x}_i, \mbox{ and } \\
  \frac{\partial \bs{H}}{\partial \lambda_s} &= 
      \bs{X}^\top \frac{\partial\bs{A}}{\partial \lambda_s} \bs{X} + 
            \frac{\partial \bs{W}}{\partial \lambda_s}.
\end{align*}
The diagonal matrices $\bs{A}$ and $\bs{W}$ have derivatives
\begin{align*}
  \left(\frac{\partial \bs{A}}{\partial \lambda_s}\right)_{ii}
    &= \threelui \times \frac{\partial u_i}{\partial \lambda_s}\quad\mbox{and} \\
  \left(\frac{\partial \bs{W}}{\partial \lambda_s}\right)_{jj}
     &= \frac{\partial \tworeg}{\partial \lambda_s}(\hat{\beta}_j) +
     \threeregj  \times \frac{\partial \hat{\beta}_j}{\partial \lambda_s},
\end{align*}
and $\utildei$ has derivatives
\begin{align*}
  \frac{\partial \utildei}{\partial u_i} &= 
    \frac{1}{1 - \twolui h_i} + \frac{\onelui \threelui h_i^2}{(1 - \twolui h_i)^2}
  \quad\mbox{and} \\
  \frac{\partial \utildei}{\partial h_i} &= \frac{\onelui}{(1 - \twolui h_i)^2}.
\end{align*}
\end{theorem}

\begin{theorem}
\label{alohessian}
The hessian of $\mathrm{ALO}_{\bs{\lambda}}$ can be computed as
\begin{align*}
  \frac{\partial^2 f}{\partial \lambda_s \partial \lambda_t} &= \frac{1}{n}\sum_{i=1}^n\left(
    \twolutildei \times \frac{\partial \utildei}{\partial \lambda_s}
                 \times \frac{\partial \utildei}{\partial \lambda_t} +
    \onelutildei \times \frac{\partial^2 \utildei}{\partial \lambda_s \partial \lambda_t}\right),
\end{align*}
where
\begin{align*}
\frac{\partial^2 \utildei}{\partial \lambda_s \partial \lambda_t} &= \!\begin{multlined}[t]
   \frac{\partial \utildei}{\partial u_i} \times
          \frac{\partial^2 u_i}{\partial \lambda_s \partial \lambda_t} +
    \frac{\partial^2 \utildei}{\partial u_i^2}\times
      \frac{\partial u_i}{\partial \lambda_s}\times\frac{\partial u_i}{\partial \lambda_t} \\
  +
      \frac{\partial^2 \utildei}{\partial u_i \partial h_i} \times \left[
        \frac{\partial u_i}{\partial \lambda_s} \times \frac{\partial h_i}{\partial \lambda_t} + 
        \frac{\partial u_i}{\partial \lambda_t} \times \frac{\partial h_i}{\partial \lambda_s} 
        \right] \\
  +  \frac{\partial \utildei}{\partial h_i} \times
          \frac{\partial^2 h_i}{\partial \lambda_s \partial \lambda_t} +
      \frac{\partial^2 \utildei}{\partial h_i^2} \times
  \frac{\partial h_i}{\partial \lambda_s}\times\frac{\partial h_i}{\partial \lambda_t},\end{multlined} \\
\frac{\partial^2 u_i}{\partial \lambda_s \partial \lambda_t} &=
      \bs{x}_i^\top \frac{\partial^2 \betahat}{\partial \lambda_s \partial \lambda_t}, \\
\frac{\partial^2 \betahat}{\partial \lambda_s \partial \lambda_t}
  &= \!\begin{multlined}[t]-\bs{H}^{-1} 
  \bs{X}^\top \cdot \mathrm{vec}\left[
    \left\{
      \threelui \times \frac{\partial u_i}{\partial\lambda_s} 
                \times \frac{\partial u_i}{\partial \lambda_t}\right\}_i\right] \\
   - \bs{H}^{-1}\frac{\partial\nabla^2 \reg}{\partial \lambda_s}(\betahat)
     \frac{\partial \betahat}{\partial \lambda_t}
   - \bs{H}^{-1}\frac{\partial\nabla^2 \reg}{\partial \lambda_t}(\betahat)
     \frac{\partial \betahat}{\partial \lambda_s}
   - \bs{H}^{-1}\frac{\partial \nabla \reg}{\partial \lambda_s \partial \lambda_t} (\betahat) \\
   -  \bs{H}^{-1} \cdot \mathrm{vec}\left[\left\{
       \threeregj \times \frac{\partial \hat{\beta}_j}{\partial \lambda_s}
    \times \frac{\partial \hat{\beta}_j}{\partial \lambda_t}\right\}_j\right],\end{multlined} \\
  \frac{\partial^2 h_i}{\partial \lambda_s \partial \lambda_t} &=
     2\bs{x}_i^\top \bs{H}^{-1} \frac{\partial \bs{H}}{\partial \lambda_s}
        \bs{H}^{-1} \frac{\partial \bs{H}}{\partial \lambda_t} \bs{H}^{-1} \bs{x}_i
    -\bs{x}_i^\top \bs{H}^{-1} 
        \frac{\partial^2 \bs{H}}{\partial \lambda_s \partial \lambda_t} \bs{H}^{-1} \bs{x}_i, \mbox{ and} \\
  \frac{\partial^2 \bs{H}}{\partial \lambda_s\partial \lambda_t} &= 
    \bs{X}^\top \frac{\partial^2\bs{A}}{\partial \lambda_s \partial \lambda_t} \bs{X} +
            \frac{\partial^2 \bs{W}}{\partial \lambda_s\partial\lambda_t}.
\end{align*}
The diagonal matrices $\bs{A}$ and $\bs{W}$ have second derivatives
\begin{align*}
  \left(\frac{\partial^2\bs{A}}{\partial \lambda_s\partial\lambda_t}\right)_{ii} &= 
    \threelui\times\frac{\partial^2 u_i}{\partial\lambda_s\partial\lambda_t} +
    \fourlui
      \times\frac{\partial u_i}{\partial \lambda_s}
      \times\frac{\partial u_i}{\partial \lambda_t} \quad\mbox{and} \\
  \left(\frac{\partial^2\bs{W}}{\partial \lambda_s \partial \lambda_t}\right)_{jj} &= 
  \!\begin{multlined}[t]\frac{\partial^2\tworeg}{\partial \lambda_s\partial \lambda_t}(\hat{\beta}_j)
  + \frac{\partial \threereg}{\partial \lambda_s}(\hat{\beta}_j)
        \times \frac{\partial\hat{\beta}_j}{\partial \lambda_t}
  + \frac{\partial \threereg}{\partial \lambda_t}(\hat{\beta}_j)
        \times\frac{\partial \hat{\beta}_j}{\partial \lambda_s} \\
   + 
    \threeregj
        \times\frac{\partial^2 \hat{\beta}_j}{\partial \lambda_s\partial \lambda_t} +
    \fourregj
        \times\frac{\partial \hat{\beta}_j}{\partial \lambda_s}
    \times\frac{\partial \hat{\beta}_j}{\partial \lambda_t},\end{multlined}
\end{align*}
and $\utildei$ has second derivatives
\begin{align*}
  \frac{\partial^2\utildei}{\partial u_i^2} &= 
    \frac{\threelui h_i}{(\ldivexpr)^2} + \frac{(\twolui \threelui + \onelui \fourlui)h_i^2}{(\ldivexpr)^2} +
        \frac{2 \onelui \threelui^2 h_i^3}{(\ldivexpr)^3}, \\
  \frac{\partial^2\utildei}{\partial u_i\partial h_i} &= 
      \frac{\twolui}{(\ldivexpr)^2} + \frac{2 \onelui \threelui h_i}{(\ldivexpr)^3}, \mbox{ and} \\
  \frac{\partial^2\utildei}{\partial h_i^2} &= \frac{2 \onelui \twolui}{(\ldivexpr)^3}.
\end{align*}
\end{theorem}

\subsection{Computational Complexity}
Let $q$ denote the number of hyperparameters. 
How best to proceed with the computations will depend on which is greater $n$ or $p$.  We assume first
that $n > p$. $\bs{H}$ and its Cholesky factorization $\bs{L}$ can be computed in 
$\bigo{p^2 n}$ operations. Let $\bs{h}$ denote the vector of $h_i$ values.
The complexity of computing the ALO value, gradient, and 
hessian is dominated by the cost of evaluating $\bs{h}$ and its derivatives. Using the formula
\begin{align*}
  h_i = \|\bs{L}^{-1} \bs{x}_i\|^2,
\end{align*}
$\bs{h}$ can be computed with $\bigo{p^2 n}$ operations. For the derivatives of
$\bs{h}$, we first compute the matrices $\frac{\partial \bs{H}}{\partial \lambda_s}$, which can be done
in $\bigo{p^2 q n}$ operations. The derivatives can then, also, be computed with $\bigo{p^2 q n}$
operations using the formulas
\begin{align*}
  \bs{t}_i = {\bs{L}^{-1}}^{\top} \bs{L}^{-1} \bs{x}_i
  \quad\mbox{and}\quad
  \frac{\partial h_i}{\partial \lambda_s} = 
      \bs{t}_i^\top \frac{\partial \bs{H}}{\partial \lambda_s} \bs{t}_i.
\end{align*}
For the second derivatives of $\bs{h}$, we, similarly, first compute the matrices 
$\frac{\partial^2 \bs{H}}{\partial \lambda_s \partial \lambda_t}$, which can be done in $\bigo{p^2q^2n}$ 
operations, and then compute
\begin{align*}
  \bs{r}_{si} = \bs{L}^{-1} \frac{\partial \bs{H}}{\partial \lambda_s} \bs{H}^{-1} \bs{x}_i 
                    \quad\mbox{and}\quad
  \frac{\partial^2 h_i}{\partial \lambda_s \partial \lambda_t} = 
    \bs{r}_{si}^\top \bs{r}_{ti} 
  - \bs{t}_i^\top \frac{\partial^2 \bs{H}}{\partial \lambda_s \partial \lambda_t} \bs{t}_i
\end{align*}
with $\bigo{p^2 q^2 n}$ operations.

When $p > n$ and $\bs{W}$ is nonsingular, we can achieve better complexity by evaluating the equations in a different order.
We apply the matrix inversion lemma to 
$\bs{H}^{-1} = \left[\bs{X}^\top \bs{A} \bs{X} + \bs{W}\right]^{-1}$ to obtain
\begin{align*}
  \bs{H}^{-1} = \bs{W}^{-1} + \bs{W}^{-1}\bs{X}^\top
  \left[\bs{A}^{-1} + \bs{X} \bs{W}^{-1} \bs{X}^\top\right]^{-1}
  \bs{X} \bs{W}^{-1}.
\end{align*}
Computing the matrix $\bs{A}^{-1} + \bs{X} \bs{W}^{-1} \bs{X}^\top$ and its Cholesky factorization 
can be done in $\bigo{n^2 p}$ operations, and a product $\bs{H}^{-1} b$ can be done in $\bigo{np}$
operations. Applying the same approach, where we avoid explicitly computing the $p$-by-$p$ matrices,
we can compute the gradient and hessian in $\bigo{n^2 q p}$ and $\bigo{n^2 q^2 p}$ operations, 
respectively. If $\bs{W}$ is singular but only has $k$ zero diagonal entries, where $k \ll p$, 
we can combine this approach with block matrix inversion and still achieve more efficient formulas.

\section{Examples}
By customizing $\ell_i$ and $R_{\lambda}$, we can adopt Theorem~\ref{alogradient} and
Theorem~\ref{alohessian} to a broad range of models. Putting $\ell_i(u) \triangleq (y_i - u)^2$ gives us
regularized least squares; putting $\ell_i(u) \triangleq \log\left[1 + \exp(-y_i u)\right]$ gives
us regularized logistic regression. If we define
\begin{align*}
  R_{\lambda}(\bs{\beta}) \triangleq \lambda^2 \left\| \bs{\beta} \right\|^2,
\end{align*}
we have standard ridge regularization. Defining
\begin{align*}
  R_{\bs{\lambda}}(\bs{\beta}) \triangleq \sum_{j=1}^p \lambda_{g_j}^2 \lvert \beta_j \rvert^2
\end{align*}
allows us to use different regularization strengths for different groups of variables, and putting
\begin{align*}
  R_{\bs{\lambda}}(\bs{\beta}) \triangleq \lambda_1^2 \sum_{j=1}^p \lvert \beta_j \rvert^{1 + \lambda_2^2}
\end{align*}
gives us bridge regularization \citep{fu:98}. We present versions of Theorem~\ref{alogradient} and
Theorem~\ref{alohessian} for the specific cases of ridge regression and logistic regression with
ridge regularization.
\subsection{Ridge Regression}
Because Equation~\ref{subobj} is a quadratic for ridge regression, the Newton approximation step in
ALO is exact, $\bs{\tilde{\beta}}_{/i} = \betahat_{/i}$, and ALO and LO are equivalent. Before
stating theorems for the LO gradient and hessian, we first introduce more familiar notation. Put
\begin{align*}
  \hat{y}_i \triangleq u_i, \quad \hat{y}_{/i} \triangleq \tilde{u}_{/i}, \quad
  \varepsilon_i \triangleq y_i - \hat{y}_i, \quad\mbox{and } 
  \varepsilon_{/i} \triangleq y_i - \hat{y}_{/i}.
\end{align*}
$\hat{y}_i$ and $\hat{y}_{/i}$ denote the $i$\ts{th} in-sample and out-of-sample prediction, 
respectively; and $\varepsilon_i$ and $\varepsilon_{/i}$ denote the $i$\ts{th} in-sample and
out-of-sample prediction error, respectively. Using this notation,
\begin{align*}
  \mathrm{LO}_{\bs{\lambda}} = \frac{1}{n} \sum_{i=1}^n \varepsilon_{/i}^2,
\end{align*}
where
\begin{align*}
  \varepsilon_{/i} = \frac{\varepsilon_i}{1 - 2 h_i}\quad\mbox{and}\quad
  h_i = \frac{1}{2} \bs{x}_i^\top \left(\bs{X}^\top \bs{X} + \lambda^2 \bs{I}\right)^{-1} \bs{x}_i.
\end{align*}
\begin{corollary}
For ridge regression, the gradient of $\mathrm{LO}_\lambda$ can be computed as
\begin{align*}
\frac{\partial f}{\partial \lambda} &= \frac{1}{n}\sum_{i=1}^n\left(
      -2 \varepsilon_{/i} \times \frac{\partial \hat{y}_{/i}}{\partial \lambda}\right),
\end{align*}
where
\begin{align*}
  \frac{\partial \hat{y}_{/i}}{\partial \lambda} &=
    \frac{\partial \hat{y}_{/i}}{\partial \hat{y}_i} \times \frac{\partial \hat{y}_i}{\partial \lambda} +
    \frac{\partial \hat{y}_{/i}}{\partial h_i} \times \frac{\partial h_i}{\partial \lambda}, \\
  \frac{\partial \hat{y}_i}{\partial \lambda} &= 
      -2 \lambda \bs{x}_i^\top \left(\bs{X}^\top \bs{X} + \lambda^2 \bs{I}\right)^{-1} \betahat, \mbox{ and}\\
  \frac{\partial h_i}{\partial \lambda} &= 
      -\lambda \bs{x}_i^\top \left(\bs{X}^\top \bs{X} + \lambda^2 \bs{I}\right)^{-2} \bs{x}_i,
\end{align*}
and $\hat{y}_{/i}$ has derivatives
\begin{align*}
  \frac{\partial \hat{y}_{/i}}{\partial \hat{y}_i} = \frac{1}{1 - 2 h_i}\quad\mbox{and}\quad
  \frac{\partial \hat{y}_{/i}}{\partial h_i} = \frac{-2 \varepsilon_i}{(1 - 2 h_i)^2}.
\end{align*}
\end{corollary}

\begin{corollary}
For ridge regression, the hessian of $\mathrm{LO}_\lambda$ can be computed as
\begin{align*}
\frac{\partial^2 f}{\partial \lambda^2} &= \frac{1}{n}\sum_{i=1}^n\left[
    2 \left(\frac{\partial \hat{y}_{/i}}{\partial \lambda}\right)^2 -
    2 \varepsilon_{/i} \times \frac{\partial^2 \hat{y}_{/i}}{\partial \lambda^2}\right],
\end{align*}
where
\begin{align*}
\frac{\partial^2 \hat{y}_{/i}}{\partial \lambda^2} &= \!\begin{multlined}[t]
   \frac{\partial \hat{y}_{/i}}{\partial \hat{y}_i} \times
          \frac{\partial^2 \hat{y}_i}{\partial \lambda^2} 
  +
      2 \frac{\partial^2 \hat{y}_{/i}}{\partial \hat{y}_i \partial h_i} \times 
        \frac{\partial \hat{y}_i}{\partial \lambda} \times \frac{\partial h_i}{\partial \lambda} \\
  +  \frac{\partial \hat{y}_{/i}}{\partial h_i} \times
          \frac{\partial^2 h_i}{\partial \lambda^2} +
      \frac{\partial^2 \hat{y}_{/i}}{\partial h_i^2} \times
  \left(\frac{\partial h_i}{\partial \lambda}\right)^2, \end{multlined}\\
\frac{\partial^2 \hat{y}_i}{\partial \lambda^2} &=
    8 \lambda^2 \bs{x}_i^\top \left(\bs{X}^\top \bs{X} + \lambda^2 \bs{I}\right)^{-2} \betahat -
    2 \bs{x}^\top\left(\bs{X}^\top \bs{X} + \lambda^2\bs{I}\right)^{-1} \betahat, \mbox{ and} \\
  \frac{\partial^2 h_i}{\partial \lambda^2} &=
    4 \lambda^2 \bs{x}_i^\top \left(\bs{X}^\top \bs{X} + \lambda^2\bs{I}\right)^{-3} \bs{x}_i -
    \bs{x}_i^\top \left(\bs{X}^\top \bs{X} + \lambda^2 \bs{I}\right)^{-2} \bs{x}_i,
\end{align*}
and $\hat{y}_{/i}$ has second derivatives
\begin{align*}
  \frac{\partial^2\hat{y}_{/i}}{\partial \hat{y}_i\partial h_i} = 
      \frac{2}{(1 - 2h_i)^2} \quad\mbox{and}\quad
  \frac{\partial^2\hat{y}_{/i}}{\partial h_i^2} = \frac{-8 \varepsilon_i}{(1 - 2h_i)^3}.
\end{align*}
\end{corollary}
\subsection{Ridge Regularized Logistic Regression}
For ridge regularized logistic regression, we have
\begin{align*}
  \bs{H} &= \bs{X}^\top \bs{A} \bs{X} + 2\lambda^2\bs{I}, \\
  \ell_i(u) &\triangleq \log\left[1 + \exp(-y_i u)\right], \\
  \onelui &= \frac{-y_i}{1 + \exp(y_i u)}, \\
  \twolui &= qp, \\
  \threelu &= q p \left(q - p\right), \mbox{ and} \\
  \fourlu &= q p \left(q^2 + p^2\right) - 4 q^2 p^2,
\end{align*}
with $p = \left(1 + \exp(-u)\right)^{-1}$ and $q = 1 - p$.
\begin{corollary}
For ridge regularized logistic regression, the gradient of $\mathrm{ALO}_\lambda$ can be computed as
\begin{align*}
\frac{\partial f}{\partial \lambda} &= \frac{1}{n}\sum_{i=1}^n\left(
      \onelutildei \times \frac{\partial \utildei}{\partial \lambda}\right),
\end{align*}
where
\begin{align*}
  \frac{\partial \utildei}{\partial \lambda} &=
    \frac{\partial \utildei}{\partial u_i} \times \frac{\partial u_i}{\partial \lambda} +
    \frac{\partial \utildei}{\partial h_i} \times \frac{\partial h_i}{\partial \lambda}, \\
  \frac{\partial u_i}{\partial \lambda} &= -4 \lambda \bs{x}_i^\top \bs{H}^{-1} \betahat, \\
  \frac{\partial h_i}{\partial \lambda} &= 
      -\bs{x}_i^\top \bs{H}^{-1} 
            \frac{\partial \bs{H}}{\partial \lambda} \bs{H}^{-1} \bs{x}_i, \mbox{ and } \\
  \frac{\partial \bs{H}}{\partial \lambda} &= 
      \bs{X}^\top \frac{\partial\bs{A}}{\partial \lambda} \bs{X} + 4 \lambda \bs{I}.
\end{align*}
The diagonal matrix $\bs{A}$ has derivative
\begin{align*}
  \left(\frac{\partial \bs{A}}{\partial \lambda}\right)_{ii}
    &= \threelui \times \frac{\partial u_i}{\partial \lambda},
\end{align*}
and $\utildei$ has derivatives
\begin{align*}
  \frac{\partial \utildei}{\partial u_i} &= 
    \frac{1}{1 - \twolui h_i} + \frac{\onelui \threelui h_i^2}{(1 - \twolui h_i)^2}
  \quad\mbox{and} \\
  \frac{\partial \utildei}{\partial h_i} &= \frac{\onelui}{(1 - \twolui h_i)^2}.
\end{align*}
\end{corollary}

\begin{corollary}
For ridge regularized logistic regression, the hessian of $\mathrm{ALO}_\lambda$ can be computed as
\begin{align*}
  \frac{\partial^2 f}{\partial \lambda^2} &= \frac{1}{n}\sum_{i=1}^n\left[
    \twolutildei \times \left(\frac{\partial \utildei}{\partial \lambda}\right)^2 +
    \onelutildei \times \frac{\partial^2 \utildei}{\partial \lambda^2}\right],
\end{align*}
where
\begin{align*}
\frac{\partial^2 \utildei}{\partial \lambda^2} &= \!\begin{multlined}[t]
   \frac{\partial \utildei}{\partial u_i} \times
          \frac{\partial^2 u_i}{\partial \lambda^2} +
    \frac{\partial^2 \utildei}{\partial u_i^2}\times
      \left(\frac{\partial u_i}{\partial \lambda}\right)^2
  +
      2 \frac{\partial^2 \utildei}{\partial u_i \partial h_i} \times 
        \frac{\partial u_i}{\partial \lambda} \times \frac{\partial h_i}{\partial \lambda} \\
  +  \frac{\partial \utildei}{\partial h_i} \times
          \frac{\partial^2 h_i}{\partial \lambda^2} +
      \frac{\partial^2 \utildei}{\partial h_i^2} \times
  \left(\frac{\partial h_i}{\partial \lambda}\right)^2, \end{multlined}\\
\frac{\partial^2 u_i}{\partial \lambda^2} &= 
  -\bs{x}_i^\top \bs{H}^{-1} \bs{X}^\top \cdot \mathrm{vec}\left[
        \left\{
          \threelui \times \left(\frac{\partial u_i}{\partial\lambda}\right)^2
                    \right\}_i\right]
       + 32 \lambda^2 \bs{x}_i^\top \bs{H}^{-2} \betahat
       - 4 \bs{x}_i^\top \bs{H}^{-1} \betahat, \\
  \frac{\partial^2 h_i}{\partial \lambda^2} &=
     2\bs{x}_i^\top \bs{H}^{-1} \frac{\partial \bs{H}}{\partial \lambda}
        \bs{H}^{-1} \frac{\partial \bs{H}}{\partial \lambda} \bs{H}^{-1} \bs{x}_i
    -\bs{x}_i^\top \bs{H}^{-1} 
        \frac{\partial^2 \bs{H}}{\partial \lambda^2} \bs{H}^{-1} \bs{x}_i, \mbox{ and} \\
  \frac{\partial^2 \bs{H}}{\partial \lambda^2} &= 
    \bs{X}^\top \frac{\partial^2\bs{A}}{\partial \lambda^2} \bs{X} + 4\bs{I}.
\end{align*}
The diagonal matrix $\bs{A}$ has second derivative
\begin{align*}
  \left(\frac{\partial^2\bs{A}}{\partial \lambda^2}\right)_{ii} &= 
    \threelui\times\frac{\partial^2 u_i}{\partial\lambda^2} +
    \fourlui
      \times\left(\frac{\partial u_i}{\partial \lambda}\right)^2,
\end{align*}
and $\utildei$ has second derivatives
\begin{align*}
  \frac{\partial^2\utildei}{\partial u_i^2} &= 
    \frac{\threelui h_i}{(\ldivexpr)^2} + \frac{(\twolui \threelui + \onelui \fourlui)h_i^2}{(\ldivexpr)^2} +
        \frac{2 \onelui \threelui^2 h_i^3}{(\ldivexpr)^3}, \\
  \frac{\partial^2\utildei}{\partial u_i\partial h_i} &= 
      \frac{\twolui}{(\ldivexpr)^2} + \frac{2 \onelui \threelui h_i}{(\ldivexpr)^3}, \mbox{ and} \\
  \frac{\partial^2\utildei}{\partial h_i^2} &= \frac{2 \onelui \twolui}{(\ldivexpr)^3}.
\end{align*}
\end{corollary}

\section{Numerical Experiments}
We run experiments designed around these lines of inquiry:
\begin{enumerate}
  \item What do the derivatives of ALO look like?
  \item How do hyperparameters found by ALO optimization compare to those found by grid search?
  \item What is the cost of ALO optimization?
  \item Can we use ALO optimization to fit models with multiple hyperparameters that lead
    to better performance on out-of-sample predictions?
  \item How closely do the ALO derivatives match up with finite difference approximations?
\end{enumerate}
To that end, we fit ridge regression and regularized logistic regression models to real-world sample
data sets. Table~\ref{tab:datasets} catalogs the data sets used, and we provide brief summaries 
below.
\begin{table}
  \centering
  \begin{tabularx}{0.7\linewidth}{@{}XXrr@{}}\toprule
    Data Set & Task & n & p\\
    \midrule
      \dsname{Breast Cancer} & classification & 569 & 30 \\
      \dsname{Cleveland Heart} & classification & 297 & 22 \\
      \dsname{Pollution} & regression & 60 & 15 \\
      \dsname{Arcene} & classification & 200 & 10000 \\
      \dsname{Gisette} & classification & 7000 & 5000 \\
      \bottomrule
  \end{tabularx}
  \caption{Sample data sets used in experiments}
  \label{tab:datasets}
\end{table}
\begin{itemize}
  \item[] \dsname{Breast Cancer} is a binary classification data set where the objective is to predict 
    whether breast mass is malignant from characteristics of cell nuclei. 
 \item[] \dsname{Cleveland Heart} is a binary classification data set where the objective is to 
    detect the presence of heart disease. The data set uses 13 features, but 5 are categorical. 
    We obtain 22 features after transforming the 
    categorical features to indicator variables, and we obtain 297 entries after dropping any 
    entries with missing features. 
  \item[] \dsname{Pollution} is a regression data set where the objective is to predict the
    mortality rate of metropolitan areas from environmental and socioeconomic variables 
    \citep{mcdonald:73}.
  \item[] \dsname{Arcene} is a binary classification data set where the objective is to distinguish 
      cancer versus normal patterns from mass-spectrometric data.  We combine the \dsname{Arcene} 
      training and validation data sets to get a data set with 200 entries. 
  \item[] \dsname{Gisette} is a binary classification data set where the objective is to predict 
      whether a handwritten digit is a 4 or a 9. The data set includes features derived from a 
      28x28 pixel digit image and non-predictive probe features.  For \dsname{Gisette}, we again 
      combine the training and validation data sets to get 7000 entries. 
\end{itemize}
We preprocess all data sets used for training so that features have zero mean and unit standard 
deviation when not constant, and we fit all models using an unregularized intercept variable.
\subsection{Visualizing ALO and its Derivatives}
\begin{figure}
     \centering
     \begin{subfigure}[b]{0.325\textwidth}
         \centering
         \includegraphics[width=\textwidth]{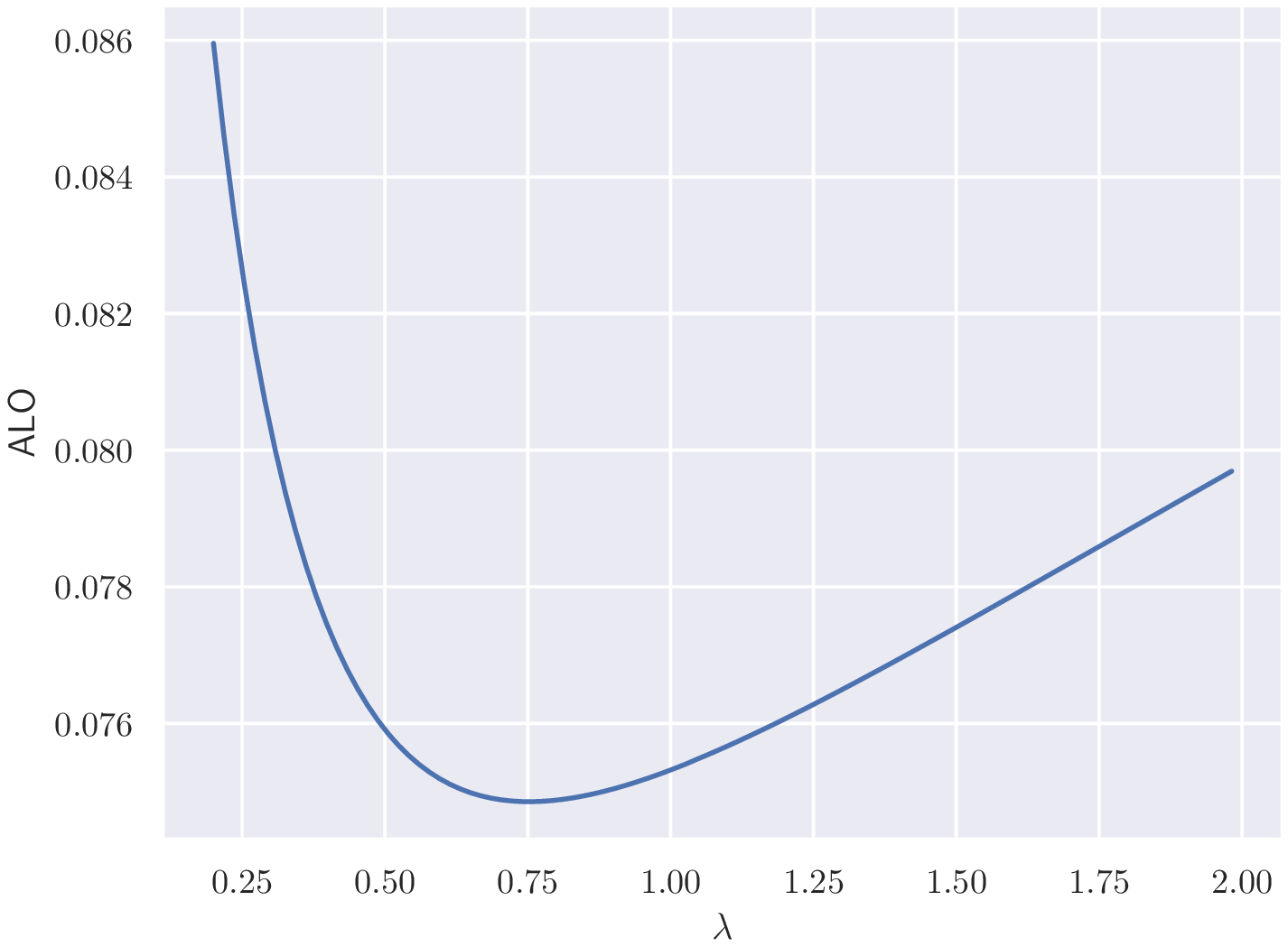}
         \caption{\dsname{ALO}}
         \label{fig:breast_cancer_value}
     \end{subfigure}
     \hfill
     \begin{subfigure}[b]{0.325\textwidth}
         \centering
         \includegraphics[width=\textwidth]{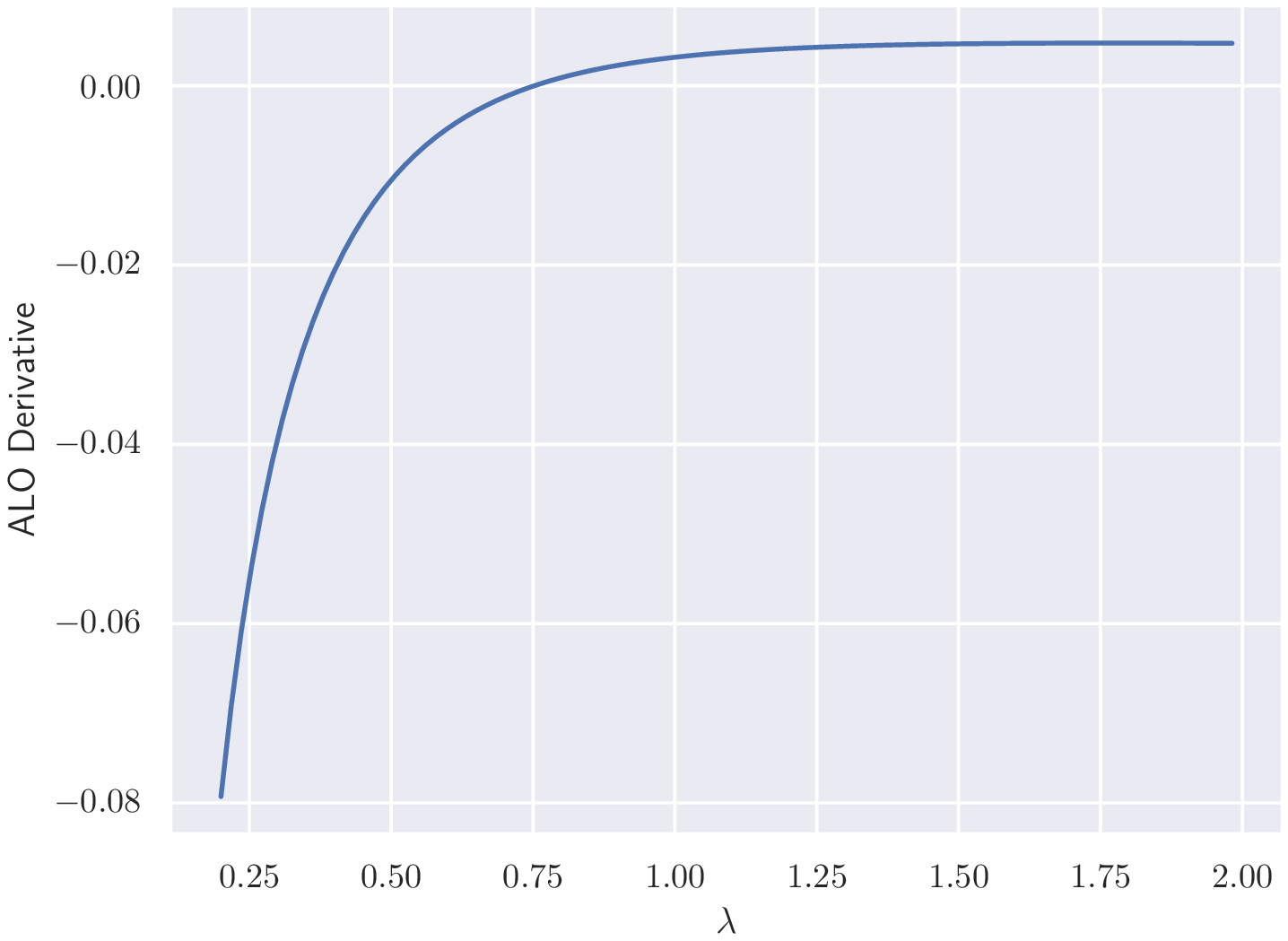}
         \caption{\dsname{ALO Derivative}}
         \label{fig:breast_cancer_gradient}
     \end{subfigure}
     \hfill
     \begin{subfigure}[b]{0.325\textwidth}
         \centering
         \includegraphics[width=\textwidth]{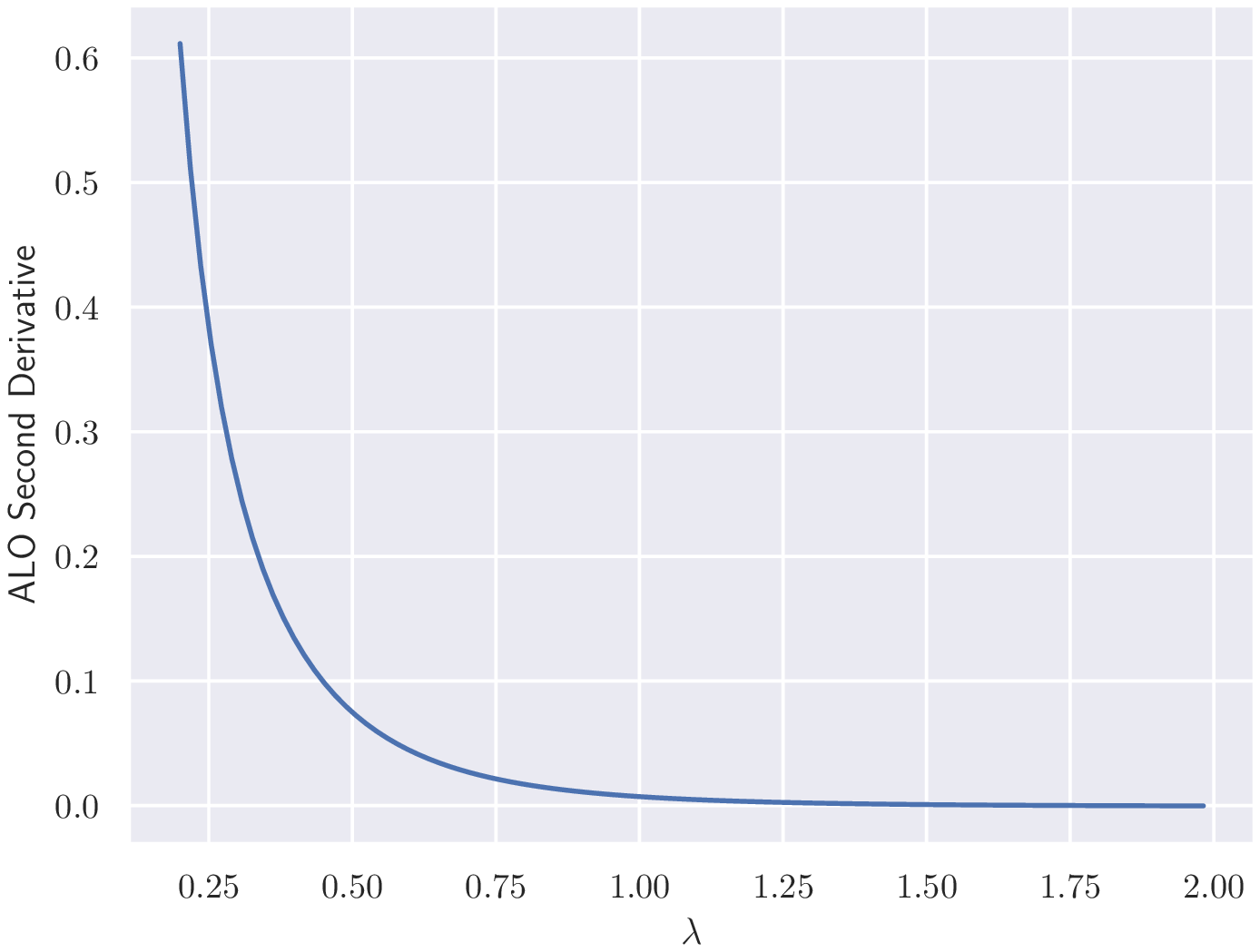}
         \caption{\dsname{ALO Second Derivative}}
         \label{fig:breast_cancer_hessian}
     \end{subfigure}
        \caption{ALO and derivatives for regularized logistic regression on the \dsname{Breast Cancer} data set. The
        regularizer is parameterized as
           $R_\lambda(\bs{\beta}) = \lambda \left\|\bs{\beta}\right\|^2$.}
        \label{fig:breast_cancer_derivatives}
\end{figure}

\begin{figure}
     \centering
     \begin{subfigure}[b]{0.325\textwidth}
         \centering
         \includegraphics[width=\textwidth]{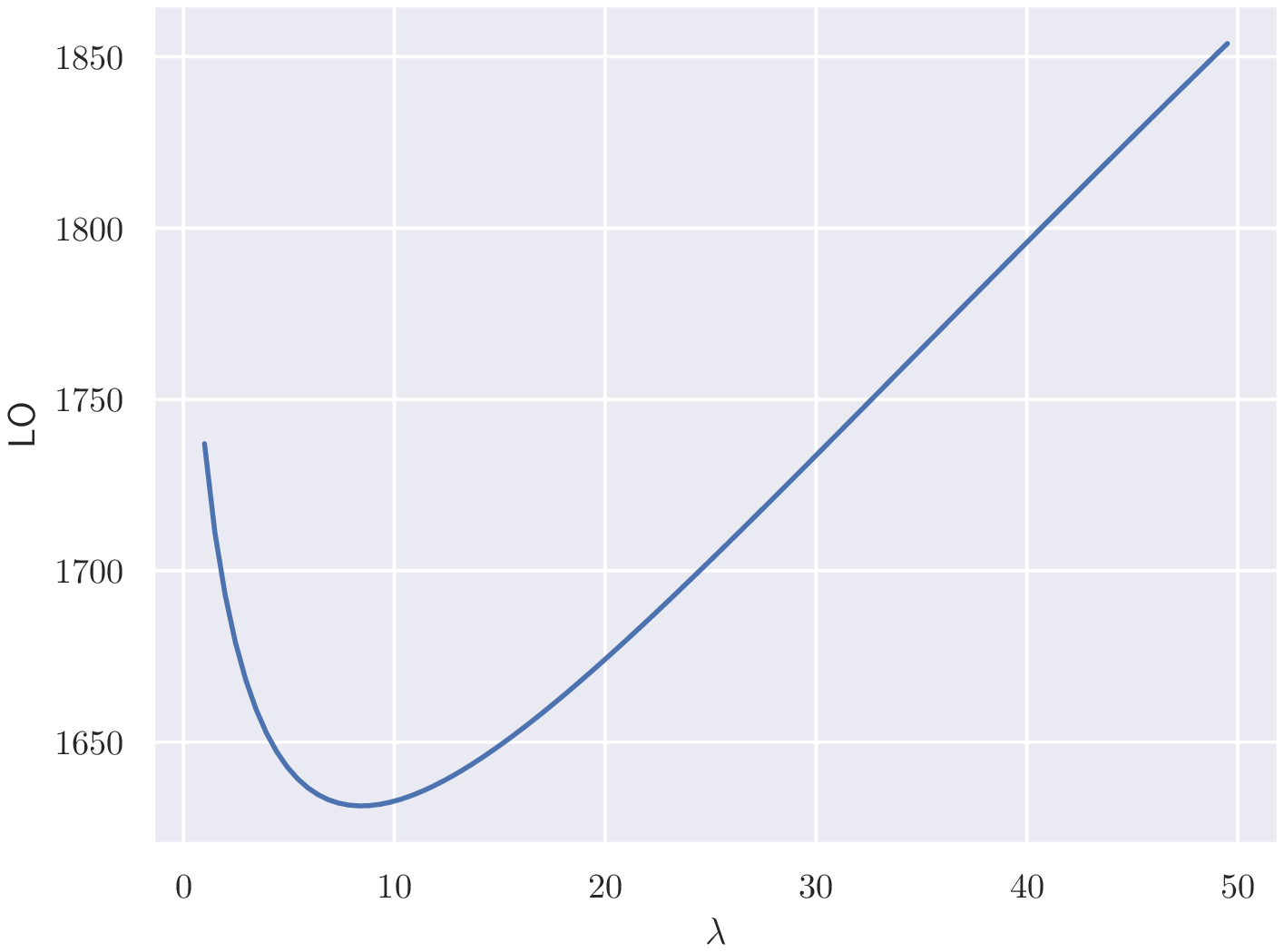}
         \caption{\dsname{LO}}
         \label{fig:pollution_value}
     \end{subfigure}
     \hfill
     \begin{subfigure}[b]{0.325\textwidth}
         \centering
         \includegraphics[width=\textwidth]{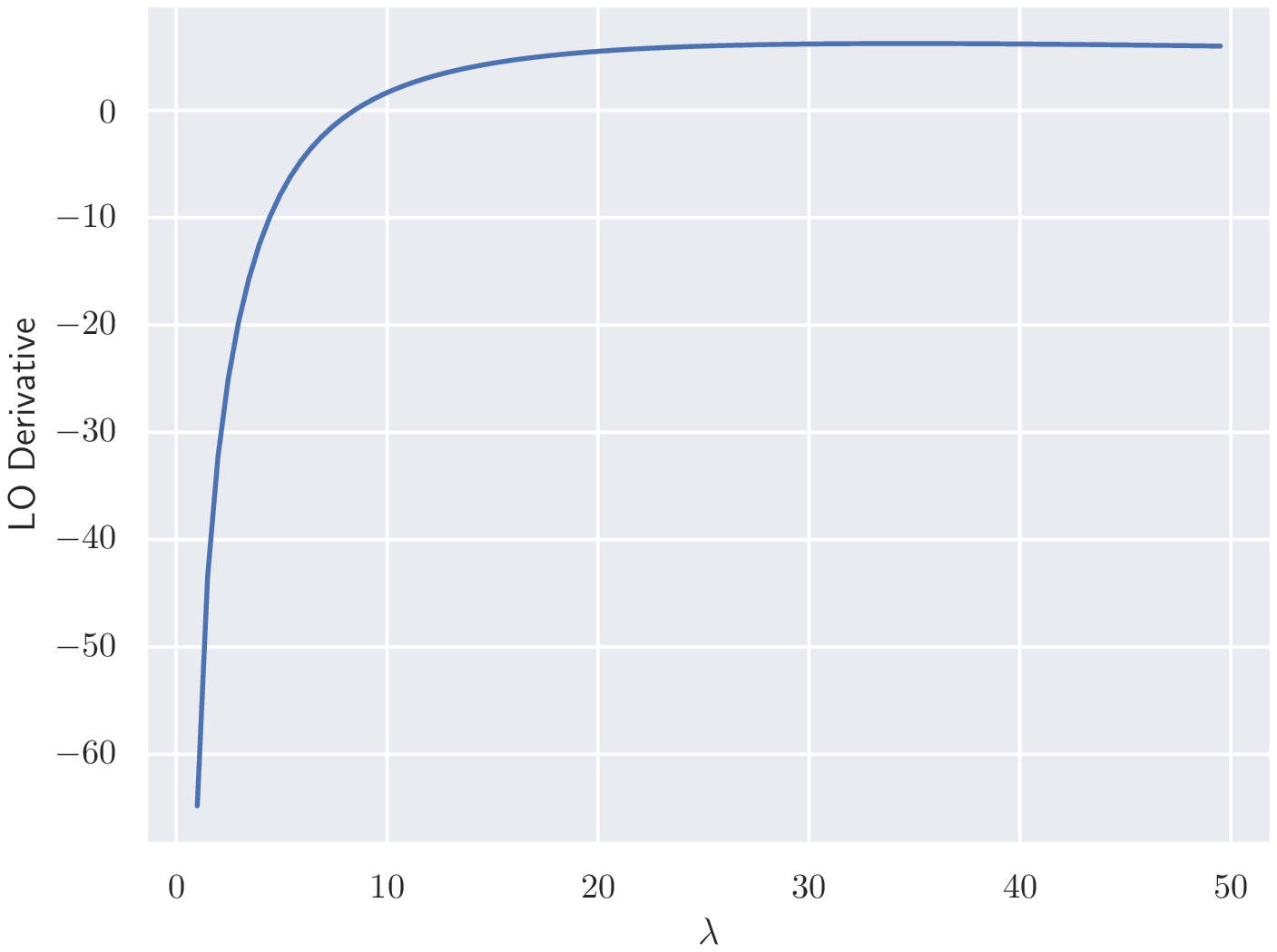}
         \caption{\dsname{LO Derivative}}
         \label{fig:pollution_gradient}
     \end{subfigure}
     \hfill
     \begin{subfigure}[b]{0.325\textwidth}
         \centering
         \includegraphics[width=\textwidth]{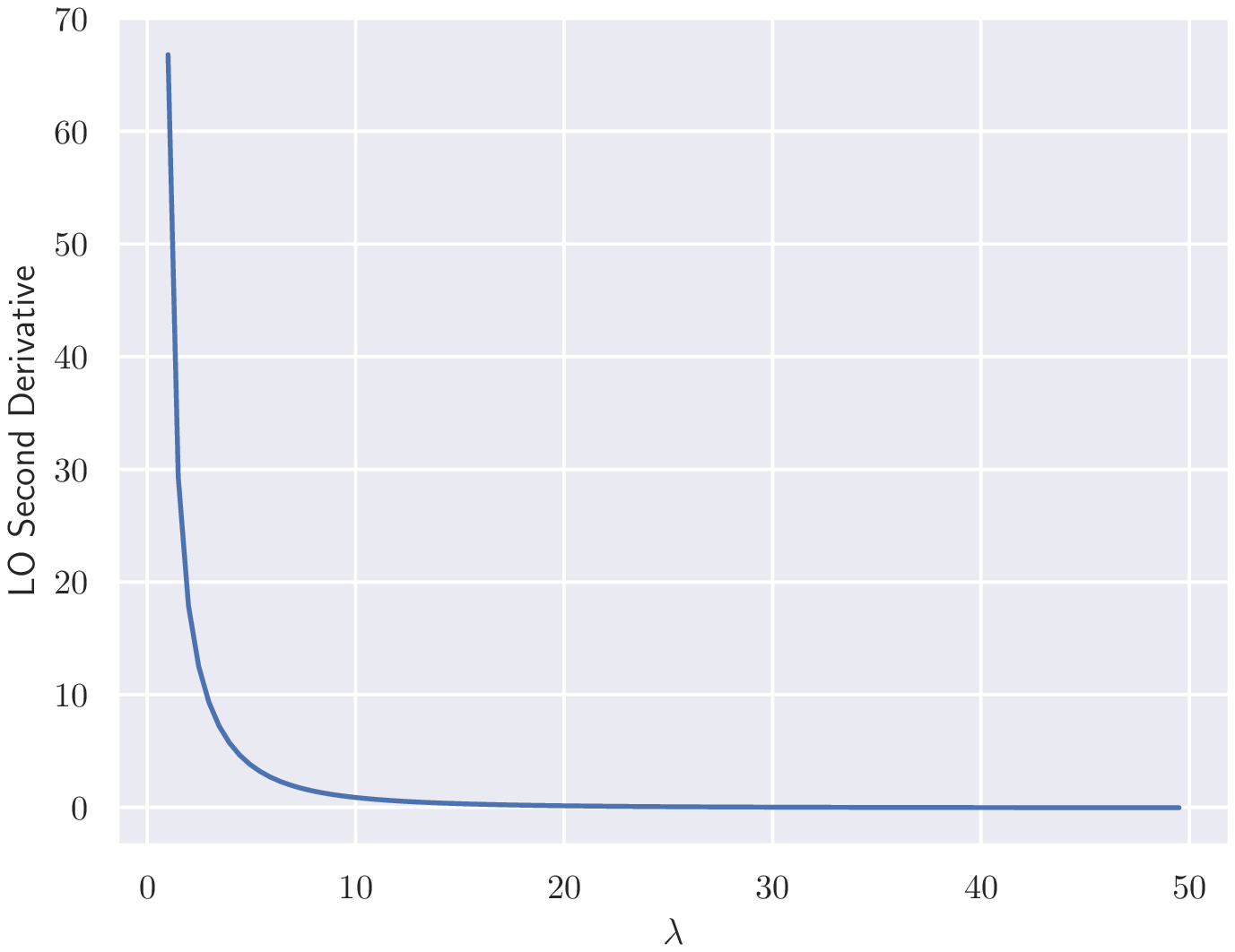}
         \caption{\dsname{LO Second Derivative}}
         \label{fig:pollution_hessian}
     \end{subfigure}
        \caption{LO and derivatives for ridge regression on the \dsname{Pollution} data set. The regularizer is 
        parameterized as
           $R_\lambda(\bs{\beta}) = \lambda \left\|\bs{\beta}\right\|^2$.}
        \label{fig:pollution_derivatives}
\end{figure}

We begin by graphing ALO and its derivatives on sample data sets. 
Figure~\ref{fig:breast_cancer_derivatives} plots ALO and its derivatives for logistic regression with the 
ridge regularizer $R_{\lambda}\left(\bs{\beta}\right) = \lambda \| \bs{\beta} \|^2$ on the 
\dsname{Breast Cancer} data set. Figure~\ref{fig:pollution_derivatives} plots LO and its derivatives
for ridge regression on the \dsname{Pollution} data set. With ridge regression, ALO and LO are equivalent.
\subsection{Comparing ALO Optimization to Grid Search}
\begin{figure}
     \centering
     \begin{subfigure}[b]{0.45\textwidth}
         \centering
         \includegraphics[width=\textwidth]{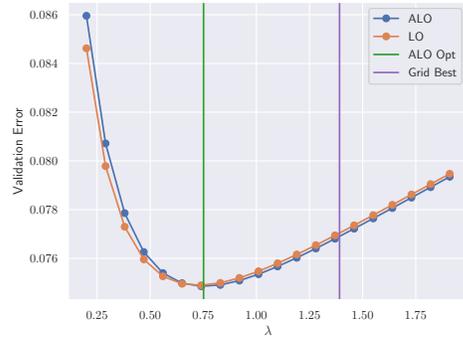}
         \caption{\dsname{Breast Cancer}}
         \label{fig:breast_cancer_opts}
     \end{subfigure}
     \hfill
     \begin{subfigure}[b]{0.45\textwidth}
         \centering
         \includegraphics[width=\textwidth]{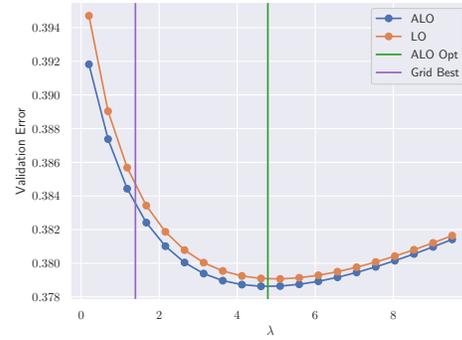}
         \caption{\dsname{Cleveland Heart}}
         \label{fig:cleveland_heart_opts}
     \end{subfigure}
     \hfill
     \begin{subfigure}[b]{0.45\textwidth}
         \centering
         \includegraphics[width=\textwidth]{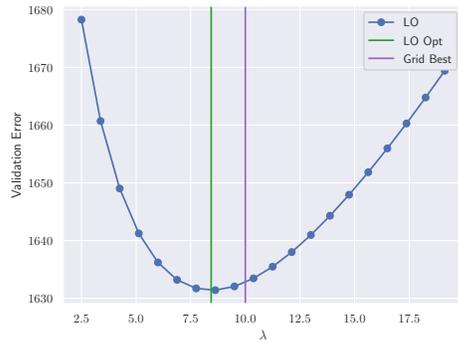}
         \caption{\dsname{Pollution}}
         \label{fig:pollution_opts}
     \end{subfigure}
     \hfill
     \begin{subfigure}[b]{0.45\textwidth}
         \centering
         \includegraphics[width=\textwidth]{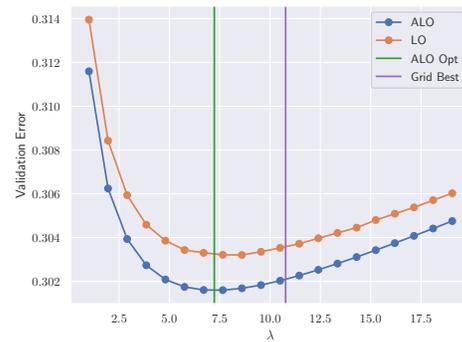}
         \caption{\dsname{Arcene}}
         \label{fig:arcene_opts}
     \end{subfigure}
        \caption{Plot of ALO and LO curves for ridge regression and regularized logistic regression on
        sample data sets. The green line shows the
        hyperparameter found by ALO optimization, and the purple line shows the hyperparameter found
        by grid search.  The regularizer is parameterized as
           $R_\lambda(\bs{\beta}) = \lambda \left\|\bs{\beta}\right\|^2$.
        }
        \label{fig:grid_compare}
\end{figure}

\begin{figure}
  \centering
  \includegraphics[width=0.75\textwidth]{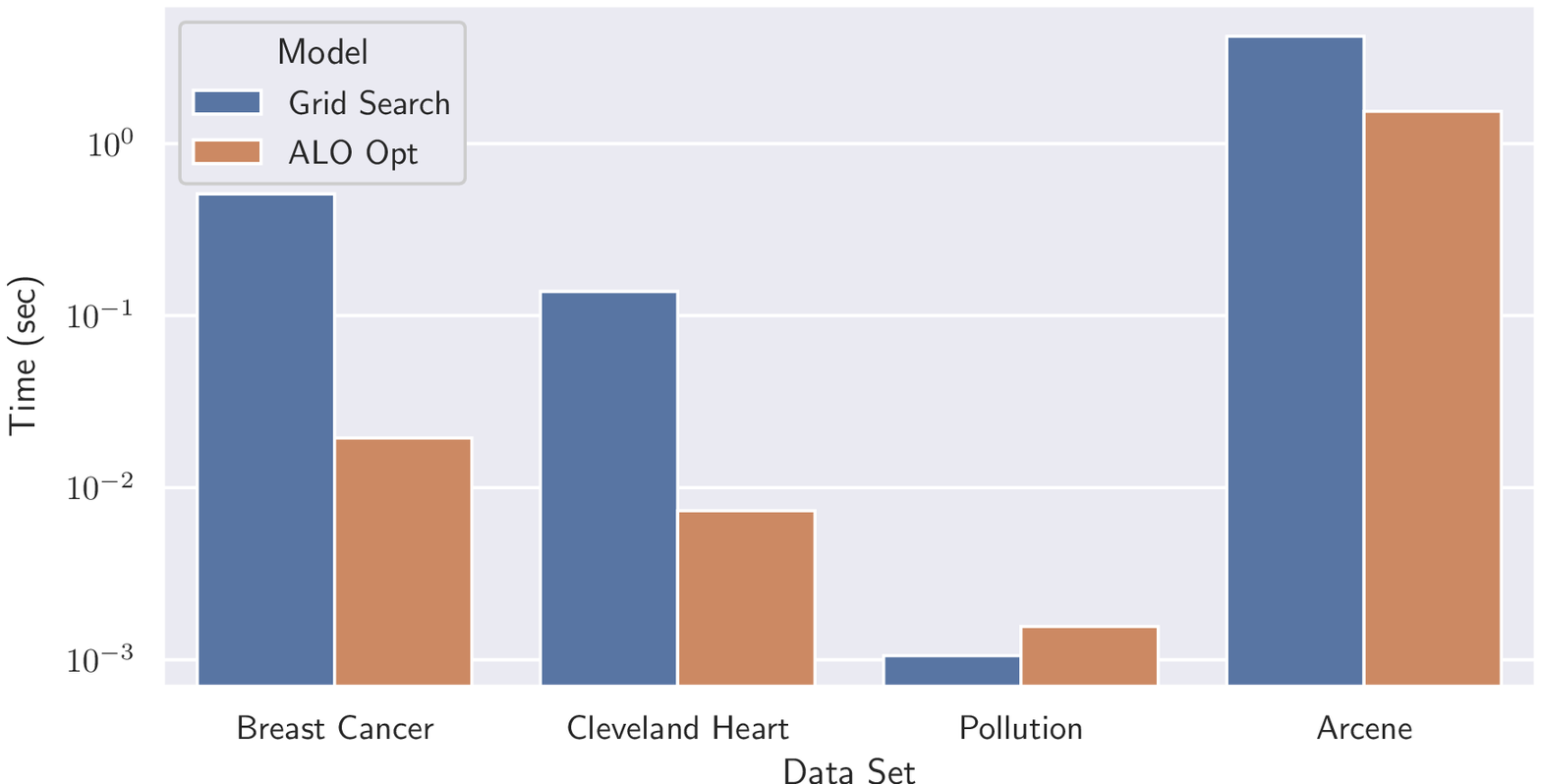}
  \caption[abc]{Benchmark showing how long ALO optimization and grid search take to find hyperparameters
  for ridge regression and regularized logistic regression on sample data sets. 
  The times are averaged over 10 runs and shown on a logarithmic scale.}
  \label{fig:benchmark}
\end{figure}

We next compare hyperparameters found by ALO optimization to those found by grid search for the ridge
regularizer. For ALO optimization, we use the Python package \package{peak-engines} available from 
\url{https://github.com/rnburn/peak-engines}; and for grid search, we use
\package{LogisticRegressionCV} and \package{RidgeCV} from \package{sklearn-0.23.1} with default 
settings. \linebreak\package{LogisticRegressionCV} defaults to run a grid search using 5-fold cross-validation
and 10 points of evaluation; \package{RidgeCV} defaults to run a grid search using LO and 3 points
of evaluation.\footnote{\package{RidgeCV}'s documentation claims that it uses Generalized 
Cross-Validation, but it's actually using LO\@. 
See \url{https://github.com/scikit-learn/scikit-learn/issues/18079}.}
Figure~\ref{fig:grid_compare} shows where the hyperparameters found by grid search and ALO
optimization lie along the the LO and ALO curves.
We additionally benchmark how long it takes ALO optimization and grid search to find hyperparameters on
the sample data sets. Figure~\ref{fig:benchmark} shows the resulting durations measured in seconds.
\subsection{Fitting Models with Multiple Hyperparameters}\label{multihyper}
To test ALO optimization with multiple hyperparameters, we fit logistic regression with
a bridge regularizer \citep{fu:98} of 
the form $\lambda_1 |\beta_j|^{\lambda_2}$, where $\lambda_2 \ge 1$, to a 5-fold cross-validation of
the \dsname{Gisette} data set.
To ensure that $r_{j,\bs{\lambda}}$ has a continuous fourth derivative, 
we interpolate the regularizer with a polynomial when $|\beta_j| < \delta$ so that
\begin{align*}
  r_{j,\bs{\lambda}}(\beta_j) \triangleq 
  \left\{
    \begin{array}{ll}
      \lambda_1 |\beta_j|^{\lambda_2}, & \mbox{if } |\beta_j| \ge \delta, \\
      \lambda_1 p_{\lambda_2}(|\beta_j|), &  \mbox{if } |\beta_j| < \delta, \\
    \end{array}
  \right.
\end{align*}
where $p_{\lambda_2}(t) = a_1 t^2 + a_2  t^4 + a_3 t^5 + a_4 t^6 + a_5 t^7$ and $\bs{a}$ is chosen so that 
$p_{\lambda_2}(\delta)$ matches $\delta^{\lambda_2}$ up to the fourth derivative. For our experiment, we
use $\delta = 0.01$. For comparison, we also fit logistic regression with ridge regularization.
Table~\ref{tab:gisette_cv} shows the hyperparameters found for bridge regularization and ridge
regularization and compares their performance on each cross-validation fold.
\begin{table}[]
  \centering
  \begin{tabularx}{0.8\linewidth}{@{}lXllrXlr@{}}\toprule
    && \multicolumn{3}{c}{Bridge} & \phantom{abc} &
    \multicolumn{2}{c}{Ridge} \\
    \cmidrule{3-5} \cmidrule{7-8}
    Fold && $\lambda_1$ & $\lambda_2$ & Test Error && $\lambda$ & Test Error \\
    \midrule
    0 && 17.3  & 1.95 & 0.0698 && 19.1 & 0.0709 \\
    1 && 16.8 & 1.94  & 0.0581 && 18.7 & 0.0587 \\
    2 && 13.0  & 1.81 & 0.0667 && 18.3 & 0.0697 \\ 
    3 && 10.5 & 1.75  & 0.0677 && 16.5 & 0.0714 \\
    4 && 10.8 &  1.83 & 0.0639 && 14.4 & 0.0667 \\
    \midrule
    Mean &&  &  & 0.0652 && & 0.0675 \\
    \bottomrule
  \end{tabularx}
  \caption{Hyperparameters and performance of ALO optimization with logistic regression using
           bridge and ridge regularization on a 5-fold cross-validation of the \dsname{Gisette}
           data set. Test error is measured as the negative mean log-likelihood of the out-of-sample
           fold.}
  \label{tab:gisette_cv}
\end{table}
\subsection{Comparing Derivatives with Finite Difference Approximations}
While working out our equations for the ALO gradient and hessian, we made extensive use of finite 
difference approximations to validate our results. We present examples of finite difference testing
for ridge regression and logistic regression using ridge and bridge regularization. We use the following
finite difference approximation to the partial derivative of a function $g$:
\begin{align*}
  \frac{\partial g}{\partial \lambda_j}\left(\bs{\lambda}\right) \approx 
        \frac{g(\bs{\lambda} + h \bs{e}_j) - g(\bs{\lambda})}{h},
\end{align*}
where $h = 0.000001$ and $\bs{e}_j$ denotes a $p$-demensional vector with the $j$\ts{th} entry equal to $1$
and all other entries equal to $0$. We approximate second derivatives by using the exact formula for the first
derivative. Table~\ref{tab:ridge_regression_finite_difference} shows the finite difference
derivative comparisons for ridge regression on the \dsname{Pollution} data set,
Table~\ref{tab:logistic_regression_finite_difference_ridge} shows the comparisons for logistic regression
on the \dsname{Breast Cancer} data set using ridge regularization, and 
Table~\ref{tab:logistic_regression_finite_difference_bridge} shows the comparisons for logistic
regression on the \dsname{Breast Cancer} data set using bridge regularization.
\begin{table}[]
  \centering
  \begin{tabularx}{0.8\linewidth}{@{}lXrrXrr@{}}\toprule
    && \multicolumn{2}{c}{$\tfrac{\partial f}{\partial \lambda}$} & \phantom{abc} &
    \multicolumn{2}{c}{$\tfrac{\partial^2 f}{\partial \lambda^2}$} \\ 
    \cmidrule{3-4} \cmidrule{6-7} 
    $\lambda$ && Exact & Approx && Exact & Approx \\
    \midrule
    0.01 && -68.99  & -69.03 && -6879.30  & -6879.27 \\
    0.05 && -33.36 & -33.39 && -6195.24 & -6195.10 \\
    0.10 && -600.79 & -600.81 && -4371.80 & -4371.59 \\
    1.00 && -129.64 & -129.63 && 137.56 & 137.56 \\
    2.00 && -48.68 & -48.68 && 65.14 & 65.14 \\
    5.00 && 59.95 & 59.95 && 18.15 & 18.15 \\
  \bottomrule
  \end{tabularx}
  \caption{Comparison of LO derivatives and finite difference approximations for ridge regression 
           on the \dsname{Pollution} data set using various values of $\lambda$. The 
           regularizer is parameterized as 
           $R_\lambda(\bs{\beta}) = \lambda^2 \left\|\bs{\beta}\right\|^2$.}
  \label{tab:ridge_regression_finite_difference}
\end{table}

\begin{table}[]
  \centering
  \begin{tabularx}{0.8\linewidth}{@{}lXrrXrr@{}}\toprule
    && \multicolumn{2}{c}{$\tfrac{\partial f}{\partial \lambda}$} & \phantom{abc} &
    \multicolumn{2}{c}{$\tfrac{\partial^2 f}{\partial \lambda^2}$} \\ 
    \cmidrule{3-4} \cmidrule{6-7} 
    $\lambda$ && Exact & Approx && Exact & Approx \\
    \midrule
    0.01 && -46.15  & -45.68 && 3850.21  & 3837.37 \\
    0.05 && -2.68 & -2.68 && 119.42 & 119.33 \\
    0.10 && -0.48 & -0.48 && 8.31 & 8.29 \\
    1.00 && -0.0064 & -0.0064 && 0.035 & 0.035 \\
    2.00 && 0.015 & 0.015 && 0.0015 & 0.0015 \\
    5.00 && 0.015 & 0.015 && -0.00041 & -0.00041 \\
    \bottomrule
  \end{tabularx}
  \caption{Comparison of ALO derivatives and finite difference approximations for regularized 
           logistic regression on the \dsname{Breast Cancer} data set using various values of $\lambda$. 
           The regularizer is parameterized as 
           $R_\lambda(\bs{\beta}) = \lambda^2 \left\|\bs{\beta}\right\|^2$.}
  \label{tab:logistic_regression_finite_difference_ridge}
\end{table}

\begin{landscape}
\begin{table}[]
  \begin{tabular}{@{}llrrrrcrrcrrcrrr@{}}\toprule
    && \multicolumn{2}{c}{$\tfrac{\partial f}{\partial \lambda_1}$} & \phantom{ab} &
      \multicolumn{2}{c}{$\tfrac{\partial f}{\partial \lambda_2}$} & \phantom{ab} &
      \multicolumn{2}{c}{$\tfrac{\partial^2 f}{\partial \lambda_1^2}$} & \phantom{ab} &
      \multicolumn{2}{c}{$\tfrac{\partial^2 f}{\partial \lambda_1 \partial \lambda_2}$} & \phantom{ab} &
      \multicolumn{2}{c}{$\tfrac{\partial^2 f}{\partial \lambda_2^2}$} \\ 
      \cmidrule{3-4} \cmidrule{6-7} \cmidrule{9-10} \cmidrule{12-13} \cmidrule{15-16}
      $\lambda_1$ & $\lambda_2$ & Exact & Approx && Exact & Approx && Exact & Approx &&
                                  Exact & Approx && Exact & Approx \\
      \midrule
      0.05 & 0.75 & -6.07  & -6.07 && -0.78 & -0.78 && 146.24 & 146.19 && 8.90 & 8.89 && 1.04 & 1.04 \\
      0.05 & 1.00 & -2.68  & -2.68 && -0.36 & -0.36 && 119.42 & 119.33 && 10.20 & 10.19 && 1.28 & 1.28 \\
      0.05 & 1.25 & -0.93  & -0.93 && -0.14 & -0.14 && 50.87 & 50.70 && 4.35 & 4.35 && 0.56 & 0.56 \\

      0.25 & 0.75 & -0.39  & -0.39 && -0.13 & -0.13 && -8.55 & -8.57 && -0.99 & -1.00 && 0.019 & 0.018 \\
      0.25 & 1.00 & -0.18  & -0.18 && -0.059 & -0.059 && 0.89 & 0.89 && 0.13 & 0.13 && 0.088 & 0.088 \\
      0.25 & 1.25 & -0.13  & -0.13 && -0.031 & -0.031 && 0.82 & 0.82 && 0.22 & 0.22 && 0.11 & 0.11 \\

      1.00 & 0.75 & 0.0054  & 0.0054 && -0.0077 & -0.0077 && 0.047 & 0.047 && 0.013 & 0.013 && 0.032 & 0.033 \\
      1.00 & 1.00 & 0.0064  & 0.0064 && -0.0021 & -0.0021 && 0.035 & 0.035 && 0.0021 & 0.0021 && 0.020 & 0.020 \\
      1.00 & 1.25 & 0.0039  & 0.0039 && 0.00062 & 0.00061 && -0.15 & -0.16 && -0.071 & -0.072 && -0.0065 & -0.0068 \\
      \bottomrule
    \end{tabular}
  \caption[Comparison of ALO derivatives and finite difference approximations for bridge regularized
           logistic regression]{
      Comparison of ALO derivatives and finite difference approximations for regularized 
      logistic regression on the \dsname{Breast Cancer} data set using various values of $\lambda_1$ and
      $\lambda_2$. 
      The regularizer is parameterized as \linebreak
  $r_{j,\bs{\lambda}}(\beta_j) \triangleq 
  \left\{
    \begin{array}{ll}
      \lambda_1^2 |\beta_j|^{1 + \lambda_2^2}, & \mbox{if } |\beta_j| \ge \delta, \\
      \lambda_1^2 p_{\lambda_2}(|\beta_j|), &  \mbox{if } |\beta_j| < \delta, \\
    \end{array}
  \right.$ where $p$ and $\delta$ are defined as in Section~\ref{multihyper}.
  }
  \label{tab:logistic_regression_finite_difference_bridge}
\end{table}
\end{landscape}

\section{Conclusion}
In this paper, we demonstrated how to select hyperparameters by computing the
gradient and hessian of ALO and applying a second-order optimizer to find a
local minimum. The approach is applicable to a large class of commonly used
models, including regularized logistic regression and ridge regression. We
applied ALO optimization to fit regularized models to various real-world data
sets. We found that when using a single-parameter regularizer, we were able to
find hyperparameters with better LO values than standard grid search approaches
and frequently were able to do so in less time. ALO optimization, furthermore,
scales to handle multiple hyperparameters, and we demonstrated how it could be
used to fit hyperparameters for bridge regularization.

\acks{We made use of the UCI Machine Learning Repository \url{http://archive.ics.uci.edu/ml} in our
experiments.
}

\appendix
\section*{Appendix A. Proof of Theorem~\ref{alogradient}}
To derive the derivative of $\betahat$, we observe that at the optimum of Equation \ref{subobj}
the gradient is zero:
\begin{align*}
  \sum_{i=1}^n \onelui \bs{x}_i + \nabla \reg(\betahat) = 0.
\end{align*}
Differentiating both sides of the equation gives us
\begin{align*}
  &\phantom{\Leftrightarrow} \frac{\partial}{\partial \lambda_s} 
    \left(\sum_{i=1}^n \onelui \bs{x}_i + \nabla  \reg(\betahat)\right) = 0 \nonumber\\
  &\Leftrightarrow \bs{H} \left(\frac{\partial \betahat}{\partial \lambda_s}\right) + 
      \frac{\partial \nabla \reg}{\partial \lambda_s}(\betahat) = 0 \\
  &\Leftrightarrow \frac{\partial \betahat}{\partial \lambda_s} = 
      -\bs{H}^{-1}\frac{\partial \nabla \reg}{\partial \lambda_s}(\betahat).
\end{align*}
For the derivative of $h_i$, we apply this formula for differentiating an inverse matrix:
\begin{align*}
  \frac{d \bs{E}^{-1}}{dt} = -\bs{E}^{-1} \frac{d \bs{E}}{dt} \bs{E}^{-1}.
\end{align*}
The other derivatives are derived as straightforward differentiations of their associated value 
formulas.
\appendix
\section*{Appendix B. Proof of Theorem~\ref{alohessian}}
For the second derivative of $\betahat$, we derive
\begin{align*}
  \frac{\partial^2 \betahat}{\partial \lambda_s \partial \lambda_t} &=
    \frac{\partial}{\partial \lambda_t} \left[
      -\bs{H}^{-1}
      \frac{\partial \nabla \reg}{\partial \lambda_s}(\betahat)
      \right] \\
    &= \bs{H}^{-1} \frac{\partial \bs{H}}{\partial \lambda_t}
       \bs{H}^{-1} \frac{\partial \nabla \reg}{\partial \lambda_s}(\betahat) 
       - \bs{H}^{-1} \frac{\partial}{\partial \lambda_t}
          \left(\frac{\partial \nabla \reg}{\partial\lambda_s}(\betahat)\right) \\
    &= -\bs{H}^{-1} 
          \frac{\partial \bs{H}}{\partial \lambda_t}\frac{\partial \betahat}{\partial\lambda_s} 
       -\bs{H}^{-1}
          \frac{\partial}{\partial \lambda_t}
          \left(\frac{\partial \nabla \reg}{\partial\lambda_s}(\betahat)\right).
\end{align*}
Now,
\begin{align*}
\left(\frac{\partial \bs{H}}{\partial \lambda_t}\right)\frac{\partial \betahat}{\partial\lambda_s}
  &= \left[\bs{X}^\top \frac{\partial \bs{A}}{\partial \lambda_t} \bs{X} +
  \frac{\partial \bs{W}}{\partial \lambda_t}\right]\frac{\partial \betahat}{\partial\lambda_s} \\
  &= 
  \!\begin{multlined}[t]\bs{X}^\top \cdot \mathrm{vec}\left[
    \left\{
      \threelui \times \frac{\partial u_i}{\partial\lambda_s} 
                \times \frac{\partial u_i}{\partial \lambda_t}\right\}_i\right]
  + \frac{\partial\nabla^2 \reg}{\partial \lambda_t}(\betahat)
     \frac{\partial \betahat}{\partial \lambda_s} \\
  + 
     \mathrm{vec}\left[\left\{
       \threeregj \times \frac{\partial \hat{\beta}_j}{\partial \lambda_s}
    \times \frac{\partial \hat{\beta}_j}{\partial \lambda_t}\right\}_j\right]
  \quad\mbox{and} \end{multlined}  \\
  \frac{\partial}{\partial \lambda_t} 
      \left(\frac{\partial \nabla \reg}{\partial\lambda_s}(\betahat)\right)
     &= \frac{\partial \nabla^2 \reg}{\partial \lambda_s}(\betahat) \frac{\partial \betahat}{\partial \lambda_t} +
       \frac{\partial \nabla \reg}{\partial \lambda_s \partial \lambda_t} (\betahat).
\end{align*}
Combining the equations, the second derivative of $\betahat$ becomes
\begin{multline*}
\frac{\partial^2 \betahat}{\partial \lambda_s \partial \lambda_t}
  = -\bs{H}^{-1} 
  \bs{X}^\top \cdot \mathrm{vec}\left[
    \left\{
      \threelui \times \frac{\partial u_i}{\partial\lambda_s} 
                \times \frac{\partial u_i}{\partial \lambda_t}\right\}_i\right] \\
   - \bs{H}^{-1}\frac{\partial\nabla^2 \reg}{\partial \lambda_s}(\betahat)
     \frac{\partial \betahat}{\partial \lambda_t}
   - \bs{H}^{-1}\frac{\partial\nabla^2 \reg}{\partial \lambda_t}(\betahat)
     \frac{\partial \betahat}{\partial \lambda_s}
   - \bs{H}^{-1}\frac{\partial \nabla \reg}{\partial \lambda_s \partial \lambda_t} (\betahat) \\
   -  \bs{H}^{-1} \cdot \mathrm{vec}\left[\left\{
       \threeregj \times \frac{\partial \hat{\beta}_j}{\partial \lambda_s}
                  \times \frac{\partial \hat{\beta}_j}{\partial \lambda_t}\right\}_j\right].
\end{multline*}
For the second derivative of $h_i$, we derive
\begin{align*}
  \frac{\partial^2 h_i}{\partial \lambda_s \partial \lambda_t} &=
    \frac{\partial}{\partial \lambda_t} 
    \left(-\bs{x}_i^\top \bs{H}^{-1} 
        \frac{\partial \bs{H}}{\partial \lambda_s} \bs{H}^{-1} \bs{x}_i \right) \\
    &= \!\begin{multlined}[t]\bs{x}_i^\top \bs{H}^{-1} \frac{\partial \bs{H}}{\partial \lambda_t}
        \bs{H}^{-1} \frac{\partial \bs{H}}{\partial \lambda_s} \bs{H}^{-1} \bs{x}_i
    + \bs{x}_i^\top \bs{H}^{-1} \frac{\partial \bs{H}}{\partial \lambda_s}
        \bs{H}^{-1} \frac{\partial \bs{H}}{\partial \lambda_t} \bs{H}^{-1} \bs{x}_i \\
    -\bs{x}_i^\top \bs{H}^{-1} 
    \frac{\partial^2 \bs{H}}{\partial \lambda_s \partial \lambda_t} \bs{H}^{-1} \bs{x}_i \end{multlined} \\
    &= 2\bs{x}_i^\top \bs{H}^{-1} \frac{\partial \bs{H}}{\partial \lambda_s}
        \bs{H}^{-1} \frac{\partial \bs{H}}{\partial \lambda_t} \bs{H}^{-1} \bs{x}_i
    -\bs{x}_i^\top \bs{H}^{-1} 
        \frac{\partial^2 \bs{H}}{\partial \lambda_s \partial \lambda_t} \bs{H}^{-1} \bs{x}_i, \\
  \frac{\partial^2 \bs{H}}{\partial \lambda_s\partial \lambda_t} 
  &= \frac{\partial}{\partial \lambda_t}
    \left(\bs{X}^\top \frac{\partial\bs{A}}{\partial \lambda_s} \bs{X} + 
            \frac{\partial \bs{W}}{\partial \lambda_s}\right) \\
  &= \bs{X}^\top \frac{\partial^2\bs{A}}{\partial \lambda_s \partial \lambda_t} \bs{X} +
            \frac{\partial^2 \bs{W}}{\partial \lambda_s\partial\lambda_t}, \mbox{ and} \\
  \left(\frac{\partial^2\bs{W}}{\partial \lambda_s \partial \lambda_t}\right)_{jj}
   &= \frac{\partial}{\partial \lambda_t} \left(
   \frac{\partial\tworeg}{\partial \lambda_s}(\hat{\beta}_j) +
        \threeregj  \times \frac{\partial \hat{\beta}_j}{\partial \lambda_s}
    \right)\\
  &= \!\begin{multlined}[t]\frac{\partial^2\tworeg}{\partial \lambda_s\partial \lambda_t}(\hat{\beta}_j)
  + \frac{\partial \threereg}{\partial \lambda_s}(\hat{\beta}_j)
        \times \frac{\partial\hat{\beta}_j}{\partial \lambda_t}
  + \frac{\partial \threereg}{\partial \lambda_t}(\hat{\beta}_j)
        \times\frac{\partial \hat{\beta}_j}{\partial \lambda_s} \\
    + 
    \threeregj
        \times\frac{\partial^2 \hat{\beta}_j}{\partial \lambda_s\partial \lambda_t} +
    \fourregj
        \times\frac{\partial \hat{\beta}_j}{\partial \lambda_s}
  \times\frac{\partial \hat{\beta}_j}{\partial \lambda_t}. \end{multlined}
\end{align*}
We omit the steps for the other derivatives as they are straightforward.

\vskip 0.2in
\bibliography{sample}

\end{document}